# Analyzing and modeling network travel patterns during the Ukraine invasion using crowd-sourced pervasive traffic data

**S. Travis Waller, PhD**
"Friedrich List" Faculty of Transport and Traffic Sciences
Technische Universität Dresden, Germany 01062
College of Engineering & Computer Science
The Australian National University (ANU), Canberra, Australia 0200
Email: steven_travis.waller@tu-dresden.de

**Moeid Qurashi, Corresponding Author**
"Friedrich List" Faculty of Transport and Traffic Sciences
Technische Universität Dresden, Germany 01062
Email: moeid.qurashi@tu-dresden.de

**Anna Sotnikova**
Department of Transport Technologies
Lviv Polytechnic National University, Lviv, Ukraine, 79013
Email: anna.o.sotnikova@lpnu.ua

**Lavina Karva**
"Friedrich List" Faculty of Transport and Traffic Sciences
Technische Universität Dresden, Germany 01062
Email: lavina.karva@tu-dresden.de

**Sai Chand, PhD**
Transport Research and Injury Prevention Centre (TRIPC)
Indian Institute of Technology Delhi
Hauz Khas, New Delhi 110016, India
Email: saichand.transport@gmail.com; saichand@iitd.ac.in

**ABSTRACT**
In 2022, Ukraine is suffering an invasion which has resulted in acute impacts playing out over time and geography. This paper examines the impact of the ongoing disruption on traffic behavior using analytics as well as zonal-based network models. The methodology is a data-driven approach that utilizes obtained travel-time conditions within an evolutionary algorithm framework which infers origin-destination demand values in an automated process based on traffic assignment. Because of the automation of the implementation, numerous daily models can be approximated for multiple cities. The novelty of this paper versus the previously published core methodology includes an analysis to ensure the obtained data is appropriate since some data sources were disabled due to the ongoing disruption. Further, novelty includes a direct linkage of the analysis to the timeline of disruptions to examine the interaction in a new way. Finally, specific network metrics are identified which are particularly suited for conceptualizing the impact of conflict disruptions on traffic network conditions. The ultimate aim is to establish processes, concepts and analysis to advance the broader activity of rapidly quantifying the traffic impacts of conflict scenarios.

## BACKGROUND

Transport behavior reflects the broader conditions of a society. As has been demonstrated repeatedly throughout history, traffic systems are acutely sensitive to external disruption. Covid-19 exemplified this globally with massive traffic and public transport disruptions (*1*). However, in addition to pandemic impacts, large-scale human conflict can be similarly disruptive. On February 24 2022, Russia invaded Ukraine as part of the conflict going back to 2014, which has resulted in a massive refugee crisis and global food shortage (*2*). The invasion began from different locations in the East and North of the country (including from the territory of Belarus). The territories near the frontline were under the constant shelling of artillery and aviation. Areas that are located far enough from the Russian and Belarusian border were also affected by long-distance missiles. As of July 19 2022, approximately 6 million refugees from Ukraine were recorded across Europe (*3*). There were close to 10 million border crossings from Ukraine within a period of five months from February 24, 2022. Furthermore, as of July 25 2022, more than 5000 civilians were killed. The invasion has resulted in acute impacts playing out over time and geography.

The events of the crisis have led to massive disruptions in road and rail movements. Extreme disruptions were seen on roads of the country's western borders, which refugees used to flee to neighboring countries such as Romania, Poland, and Hungary. With around 370,000 refugees at the Poland-Ukraine border trying to flee, the travel time reached up to 60 hours during the first week of the invasion. Similarly, long queues were reported in borders leading to Moldova, with a wait time of 15-30 hours (*4*). Even before the invasion, the motorways in Ukraine were in a disadvantaged condition compared to their European counterparts, as reported by (*5*). During the invasion, the major roads were extensively blocked. Disruption was further worsened by the suspension of ground travel, destruction of key bridges to slow the advance of Russian troops, and snowy roads. This led to the use of secondary roads, which are in further poor condition than the highways. The regular rail services were also suspended. Evacuation trains were operating from Kyiv and other big cities in the east of the country; most of them were travelling to Lviv or through Lviv to the western border. From Lviv, trains to Przemyśl were departing one after the other, boarding on a first-come, first-served basis (*6*).

Unfortunately, the literature on travel patterns during such human-driven large-scale (and sustained) disruptive events is scarce. On the contrary, the global disruption of the COVID-19 pandemic has received a significant degree of research attention (*1*, *7*, *8*). There have been some studies evaluating the shift in travel behavior after major disruptive events such as terrorist attacks. For example, passenger journeys on the London tube decreased significantly, and private vehicle trips increased after the July 7 bombings (*9–11*). A similar pattern of public transport avoidance was observed after the September 11 attacks in New York, USA (*12*). Nevertheless, for more limited attacks (e.g. London), the effects lasted less than four months, while the September 11 attacks had prolonged effects lasting for one to two years (*13*). Finally, there also have been a few studies evaluating the "impact" of violent demonstrations (*14*), terrorism (*15*), domestic security (*16*), political uncertainty (*17*), etc., on the tourism sector. However, studies analyzing travel patterns during large-scale conflicts or invasions appear to be exceptionally limited within the transportation science literature.

Conducting transport assessments and developing transport modelling tools for any city/country often involves surveying the existing transport system and its operations. Historically, this requires using physical apparatus and/or human resources to measure vehicle properties (volumes, speeds etc.), land use and capture demographic qualities of the study region (*18*). This is a time-consuming and costly exercise that also has accuracy limitations associated with the timing and sampling of the survey deployment (*19*). These issues are exacerbated in large-scale conflict situations where existing physical infrastructure is sometimes damaged, or human resources are not readily available to process data from the field. Having access to good quality data during these turbulent times would be invaluable as these are once-in-a-lifetime (perhaps even rarer) events. Such data could be helpful in planning for future events, designing better infrastructure for

*Waller et al.*

faster evacuation, reducing uncertainties in travel, and quantifying the level of conflict using travel measures as a surrogate.

In order to overcome these challenges, it is necessary to utilise low-resource, simple and accessible data options. Smartphone technology has revolutionised the ability to track mobility patterns and the use of transport infrastructure leading to "crowd-sourced traffic data". Such data can potentially provide the required observability and ability to track the use of transport infrastructure. Specifically, the crowd-sourced pervasive traffic data and navigation providers, such as Google and TomTom, remain mostly active and accessible. They have comprehensive spatial coverage and acceptable temporal resolution and are cost-effective compared to traditional data sources (*20*). Many recent studies have utilized such datasets for various applications such as automated transport planning and demand estimation (*18*, *20*), designing adaptive traffic signals (*21*), and congestion estimation (*22*).

Here, we employ an Automated/Rapid Transport Planning framework, which utilises observed crowd-sourced transport performance metrics to infer travel demand data, which inverts the traditional approach of defining or estimating demand to assess potential impacts across a network model (*20*). This leverages data that is easier to collect through crowd-sourced options such as travel times and speeds instead of collecting data related to trip generation and origin-destination (OD) mapping. In this study, we use TomTom Application Programming Interface (API) data to examine travel behavior in selected key Ukrainian cities for roughly five weeks following the start of the invasion. First, we evaluate the reliability of data quality. Then, we evaluate the spatial and temporal variations in link travel times and congestion for each city within the context of the Ukraine disruption timeline. And finally, we present the analysis of travel demand and traffic congestion patterns for the three main attacked cities.

## UKRAINE WAR LOCATIONS AND TIMELINE

On February 24, 2022, Russia started the full-scale war in Ukraine. The invasion has begun from different locations in the east and north of the country (including from the territory of Belarus). The territories near the frontline were under the constant shelling of artillery and aviation. Areas that are located far enough from Russian and Belarusian borders were also attacked by long-distance missiles. Every city is in different conditions depending on its geographical location, urban facilities, transport connection etc. Below we describe the characteristics of six critical Ukraine cities that are considered in this study.

**Kyiv** being the capital city, was the main target of Russia. However, due to the strong defense forces, taking control of this city was challenging. Only the suburban cities in the Northeast of Kyiv were occupied, from which evacuation was almost impossible. Given the threat, the citizens of Kyiv tried to escape from the very first day of the invasion.

**Lviv** is located in the West of Ukraine (70km from the Polish border) and has a direct train connection with Poland. From the beginning of the full-scale war, it became a huge logistic and humanitarian hub. Millions of people from the whole country were arriving in the city and staying there or continuing their trips to smaller settlements or abroad. Moreover, a lot of humanitarian help from European countries was shipped to Lviv and then distributed to other places.

**Odesa** is situated on the bank of the Black Sea. Being a port city, it plays an important role in freight distribution to and from Ukraine, especially in wheat supply.

**Dnipro** can be called a medical hub of Ukraine. A lot of injured people from the East and South of the country are first transferred here. And also, together with Zaporizhzhia, it was a city where evacuation vehicles arrived from occupied territories.

**Kharkiv** is the second largest Ukrainian city by population. It is located only 30 km from the Russian border and was also attacked from the first day of the full-scale war. Constant shelling of the city by Russian troops caused a huge wave of evacuation. As a consequence, as per some sources, more than 700,000 residents, which is about half of the city's population, fled the city (*23*).

**Mariupol** is the only large city or big barrier for Russia to create an overland corridor from Russian territory to occupied Crimea, therefore becoming one of the prime occupation targets. From March 1, 2022, the city was surrounded by Russian troops and was exposed to constant shelling. Most buildings were destroyed, and most evacuation attempts were foiled by the occupying forces. Also, from the middle of March, there was forced deportation of the population to the territory of Russia.

Table 1 enlists the timeline of key events in each of the six cities studied in the paper (a more detailed version of which is made available online (*24*)). It is formulated via a synthesis of data collected from various sources, including Ukrainian social media platforms (*25*, *26*), other trusted mass media (*27*, *28*), socio-political organizations (*29*, *30*), and official channels of communication of regional military (state) administrations. The information has been cross-checked for reliability using multiple data sources.

**Table 1 Ukraine war timeline over the study period**

| Kyiv | | |
|---|---|---|
| Ref. | Event | Date(s) |
| 1 | A series of powerful airstrikes on various objects in Kyiv | 24.2.2022 |
| 2 | Battles on Peremogy Avenue and Degtyarivska Street (west part of the city) | 25-26.2.2022 |
| 3 | Rocket attack on a residential building; Kyiv metro goes into shelter mode; passenger transportation is not carried out | 26.2.2022 |
| 3 | Curfew | 26-28.2.2022 |
| 4 | Hit on radioactive waste disposal site of the Kyiv branch of "Radon Association". | 28.2.2022 |
| 5 | Hit in the direction of the TV tower | 1.3.2022 |
| 6 | A Russian projectile hit the Lavina Mall shopping center | 14.3.2022 |
| 7 | Curfew | 15-17.3.2022 |
| 8 | Russian missile partially destroyed Retroville shopping center | 20.3.2022 |
| 9 | Deoccupation of the whole Kyiv region | 2.4.2022 |
| | | |
| **Kharkiv** | | |
| Ref. | Event | Date(s) |
| 1 | Russian troops began shelling Kharkiv | 24.2.2022 |
| 2 | Massive shelling of residential areas (thirteen times). Several Russian tanks entered Kharkiv | 26.2.2022 |
| 3 | Rocket attack on Freedom Square; regional state administration building partially destroyed; bombs, rockets and shells hit residential buildings and civilian objects. (Casualties: 23) | 1.3.2022 |
| 4 | Mass attack on residential areas in which "Northern Saltivka" micro-district was most affected (40 apartment buildings destroyed, Casualties: 34) | 3.3.2022 |
| 5 | Missile strikes on the Regional State Administration building, Assumption Cathedral, and Karazin University. Shelling of sleeping areas | 4.3.2022 |
| 6 | Russian troops tried to storm Kharkiv. Artillery shelling continued. | 15.3.2022 |
| 7 | The market "Barabashovo" and the town of Merefa were shelled, destroying a school and a cultural center (Casualties: 28) | 17.3.2022 |

| 8 | At least 50 shellings during the day. The Russian military blew up one of the gates of the Oskil reservoir dam (Casualties: 11) | 3.4.2022 |
| 9 | During the night, time-delayed landmines were scattered remotely using artillery in various districts (Casualties: 7) | 11.4.2022 |
| | | |
| **Mariupol** | | |
| **Ref.** | **Event** | **Date(s)** |
| - | Shelling of the city | 24.2.2022 (until now) |
| 1 | Tanks moved from Donetsk towards Mariupol but were destroyed by the Ukrainian army | 27.2.2022 |
| 2 | In the evening, electricity, gas, and the Internet were cut off in most areas of the city. | 28.2.2022 |
| 3 | Encirclement and blockade of the city by Russia | 1.3.2022 (until now) |
| 3 | Strikes in all areas of the city, including critical and communal infrastructure objects. Another attempt to break through the defense of Mariupol | 1.3.2022 |
| 4 | Russian troops shelled the Epicenter shopping center, the 22$^{nd}$ and 17$^{th}$ neighborhoods and a blood transfusion station | 3.3.2022 |
| 5 | The capture of Mangush and exit to the sea | 8.3.2022 |
| 6 | An airstrike destroyed a maternity hospital and a hospital in the city center | 9.3.2022 |
| 7 | The capture of Naydenivka, Lyapin, Vynogradar, Sartana | 10.3.2022 |
| 8 | The capture of Volnovakha and the eastern suburbs of Mariupol | 12.3.2022 |
| 9 | "Green corridor" for evacuation | 15-18.3.2022 |
| 10 | Airstrike on the Mariupol Theater (bomb shelter). Russian army broke through the eastern part of the city. | 16.3.2022 |
| 11 | Ukrainian military controls only half of the city, while the occupiers control 17-23 micro districts, the Left Bank, and other parts of Mariupol | 17.3.2022 |
| 12 | Battles for individual buildings and whole blocks | 23.3.2022 (until 28.03.2022 |
| | | |
| **Dnipro** | | |
| **Ref.** | **Event** | **Date(s)** |
| 1 | Three airstrikes at a kindergarten, an apartment building and a shoe factory | 11.3.2022 |
| 2 | Missile attack on the Dnipro International Airport | 15.3.2022 |
| 3 | Rocket attacks an oil depot and a facility in Novomoskovsk | 30.3.2022 |
| 4 | Rocket attack on a military unit in the Dnipropetrovsk region (Casualties: 2) | 31.3.2022 |
| 5 | Rocket attack destroying a civil infrastructure object | 2.4.2022 |
| 6 | Rockets hit an oil depot with fuel and a factory | 6.4.2022 |
| 7 | With three shellings, Russian troops struck the Sinelnykivsky and Kryvorizky districts (Casualties: 4) | 7.4.2022 |
| 8 | Multiple hits together in different corners of the region. Seven strikes on the Dnipro during the night | 10.4.2022 |
| | | |
| **Lviv** | | |
| **Ref.** | **Event** | **Date(s)** |
| 1 | Three military units attacked | 24.02.22 |
| 2 | Airstrike on the International Center for Peacekeeping and Security | 13.03.22 |
| 3 | Missile strike on Lviv State Aircraft Repair Plant | 18.03.22 |

*Waller et al.*

| 4 | Two heavy missile attacks on an oil depot and Lviv Armored Plant | 26.03.22 |
| --- | --- | --- |
| 5 | Missiles hit on car maintenance station and a military warehouse | 18.04.22 |
| | | |
| **Odesa** | | |
| Ref. | Event | Date(s) |
| 1 | Citizens' evacuation begins via trains | 27.2.2022 |
| 2 | Rocket fire damaged a gas pipeline | 1.3.2022 |
| 3 | Airstrikes on the Bilenke village and Zatoka town | 3.3.2022 |
| 4 | Russian troops struck settlements in the Odesa region and also attacked by ships | 15-16.03.2022 |
| 5 | Two ships shelled the coastal zone in the southern part of the city | 21.3.2022 |
| 6 | The Odesa oil refinery and several oil storage facilities were shelled | 3.4.2022 |
| 7 | Missile attack on infrastructure facilities | 7.4.2022 |
| 8 | Curfew | 9-11.4.2022 |

## TRAVEL TIME DATA

**Data collection**
We utilize a new developed OD estimation and network inference methodology (*20*) for 7 city road networks (6 urban networks and 1 highway network in the entire country) in Ukraine. Firstly, we extracted the OpenStreetMap road network data for each network based on the bounding box coordinates. Next, we extracted the link travel time data from TomTom for all the networks. The data collection was done in two periods, viz. February 25 to March 16 and March 25 to April 12 2022. A minimum of three different departure times were set throughout the collection period, i.e., morning at 9 am, afternoon at 1 pm, and evening at 5 pm.

**Data reliability**
Although pervasive traffic data are potentially useful in disruptive events, their reliability has never been assessed due to the scarcity of such events. For example, using averaged or historical traffic values in case no information sources are detected is a common practice for the navigation providers like Google and TomTom. Therefore, part of the collected data might only represent routine/average network conditions. Such cases could also be somewhat visible, specifically in the case study of the Ukraine invasion, with some navigation platforms temporarily turning off the live feature. Therefore, it is also essential to measure data reliability before inferring any travel demand or mobility patterns.

In this section, we assess the data quality by analyzing the frequency of link travel time updates. We assume that the travel time updates are novel only if they are unique from the other detected times. Furthermore, the analysis is conducted on the six major Ukrainian cities and the Ukraine intercity highway network. It also filters out the shorter length links, i.e., shorter than 100 meters for all cities and shorter than 500 meters for the intercity highway network, which can dilute the findings. Figure 1(a) shows the histogram of the number of links with their respective unique travel time values observed over the study period (36 days). Similarly, Figure 1(b) provides the probability density functions for all cities to provide a normalized comparison of all cities. It is evident that for all cities apart from Lviv, the links are uniquely updated between 5 to 15 times over the time course regardless of their size, while for Lviv, the network links get more frequent time updates and therefore show a more distributed probability density function. The Ukraine intercity highway network shows the least data quality, where most of the network links only contain up to 10 unique travel time values.

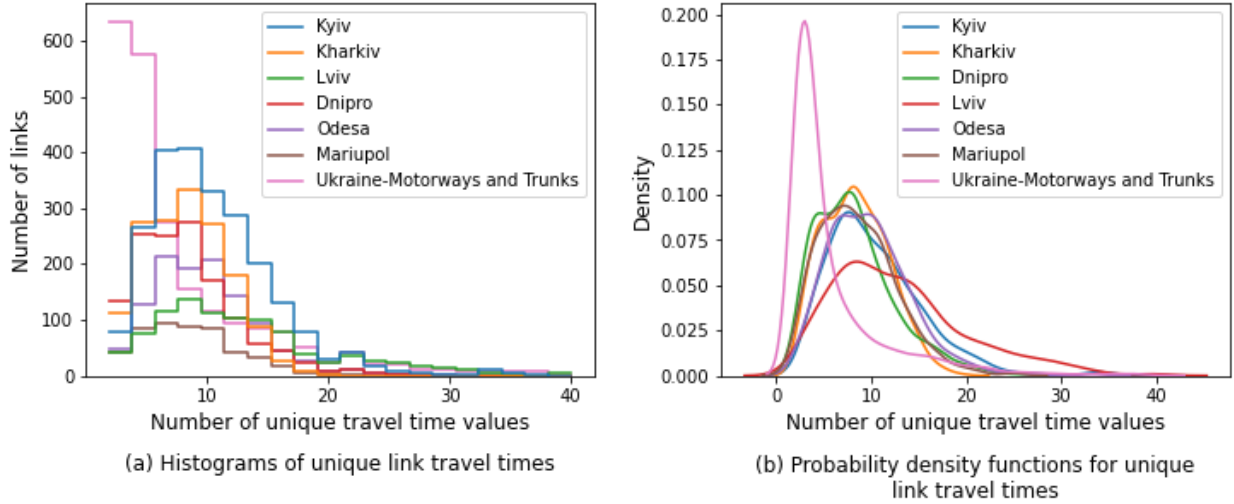

*Figure 1 TomTom data reliability plots*

## ANALYSING TRAFFIC PATTERNS FROM TRAVEL TIME DATA

Link travel times are the best source to directly observe the network traffic patterns; thus, it has been extensively used by practitioners for estimating and predicting travel demand as well as traffic states. Similarly, for disruptive events like the Russian invasion of Ukraine, link travel times are best suited to help analyze the resulting travel patterns and their relationship with different war disruptions. Therefore, in this section, we utilize the Ukraine travel time dataset i) to spatially analyze the link travel time variability averaged over the study period and ii) to temporally analyze the link travel time variability and congestion levels throughout the study period. The analysis covers the interstate highway network and six major cities of Ukraine, i.e., Kyiv, Kharkiv, Lviv, Dnipro, Odesa, and Mariupol. Note that the two mentioned temporal analyses also relate the observed network disruptions with the war events listed in the section 'Ukraine War Locations and Timeline'. As per the authors' knowledge, this is one of the first kinds of analysis of travel patterns for such major disruptive events.

**Metrics**

Several methods of calculating the travel time variability, such as standard deviation, coefficient of variance (CoV), buffer time index, planning time index, etc., have been explored in the literature. The most frequently used method of evaluating the travel time reliability is using the CoV of the travel time, which considers both the variation and the average travel time on a link (*31*). It allows demonstrating the normalised variability in travel time.

The following equations show the calculation of coefficients of variance.

$$CoV_i^{dW} = \frac{\sigma_i^{dW}}{\mu_i^{dW}} \qquad \text{(Eq. 1)}$$

$$CoV^{dW} = \frac{\sum CoV_i^{dW}}{n} \qquad \text{(Eq. 2)}$$

Where $CoV_i^{dW}$ is the coefficient of variance of link *i* during a moving window of *W* days for a departure time *d*. Similarly, $\sigma_i^{dW}$ and $\mu_i^{dW}$ denote the corresponding standard deviation and mean of travel times for link *i* during a moving window of *W* days for a departure time *d*. Note that $i \in I$, where *I* is a set of *n* network links, and $CoV^{dW}$ denotes the average network coefficient of variance.
In this study, we considered a moving window of 7 days.

Similarly, many researchers have provided varying definitions and metrics for traffic congestion that are centred on traffic characteristics such as exposure (*32*), travel time (*33*), delay (*34*), speed (*35*), volume to capacity ratio (*36*), level of service (*35*), and travel cost (*37*). Each metric has its own merits and demerits. In this study, we consider the ratio of travel time and the free-flow travel time of the link as a measure of congestion level. The following equations show the process of calculating a moving average congestion level for the entire network.

$$CI_i^{dx} = \frac{T_i^{dx}}{F_i^{dx}} \qquad (Eq.\ 3)$$

$$CI^{dx} = \frac{\sum CI_i^{dx}}{n} \qquad (Eq.\ 4)$$

Where $CI_i^{dx}$ is the congestion index of link *i* at a departure time of *d* on a specific day *x*. Similarly, $T_i^{dx}$ and $F_i^{dx}$ denote the corresponding standard deviation and mean of travel times for link *i* at a departure time of *d* on a specific day *x*. Finally, $CI^{dx}$ and *n* denote the average network congestion level and the number of links, respectively.

**Spatial analysis of link travel time variability**
Figure 2 spatially visualizes the coefficient of variance (CoV) metric of travel time for each network link over the whole study period (morning departure time). The metric is scaled using a hue scale that ranges from dark blue depicting lower variability, to dark red depicting higher variability. While analyzing the figure, it is evident that, for all cities, links in the outer city regions show much higher variability than the inner (or city center) regions. The resulting patterns can be arguably interpreted as since inner city regions are densely populated, have high shares of regular commutes, and are prone to more frequent war incidents (e.g., shelling or missile attacks or occupation attempts), they might result in lesser but consistent traffic congestions. In comparison, the higher variability and disruption in the outer region are due to factors like irregular blockage/availability and irregular collective migration patterns to and from the city affected by the war situations.

Further regarding individual cities, the three cities of Kyiv, Kharkiv, and Mariupol were under attack from the first day of the war (see Table 1), and therefore its effect is evident in Figure 2, where the hue maps of all three cities show overall higher variability in link travel times compared to other cities (showing lighter blue tone to red tone). Note that, while comparing these three cities, the city of Mariupol, which has been most affected by the Russian invasion, does show the highest variability, Kyiv, having strong defense capture, shows the least variability inside the city, and Kharkiv shows higher variability in north, south, and southeast region which contain radial links that connect the city towards Russia, Russian occupied territories, and Dnipro.

Similarly, the other three cities of Lviv, Dnipro, and Odesa show similar patterns, where most regions show the least variability while some show very high travel time variability. Since Dnipro, together with Zaporizhzhia, were the cities where evacuation vehicles arrived from occupied territories, almost all radial links for accessing the city depict high variability. Whereas Lviv being the least directly affected city through Russian attacks, show the least variability. For Lviv, although it is mentioned that it acted as the logistic and humanitarian hub with millions of people from the whole country consistently arriving in the city, the east city links depict a rather mild variability compared to Dnipro, whereas the west side links do depict very high variability.

Finally, regarding the travel patterns in Ukraine's intercity highway network, the effects of the Ukraine war are evident. All migration routes, including the westside highway network of Ukraine that connects to Lviv and Europe and the highways to Moldova, show much higher travel time variability. Similarly, the route from Dnipro to Kharkiv also depicts high variability since the city act as the destination for evacuation vehicles. Moreover, the highways connecting Kyiv from north (Gomel) and northwest, and the highway connecting Kharkiv to north (Russia) were under the Russian occupation and also show high variability by Russian military movements. Whereas the highway connecting Kyiv to western region was blocked by Russian by destroying a large bridge en-route. Likewise, other access routes to Kyiv and Kharkiv also show a mix of high and low variability links (can be due to blockages/destroyed bridges, limited data observability, etc.), depicting varying traffic congestion patterns.

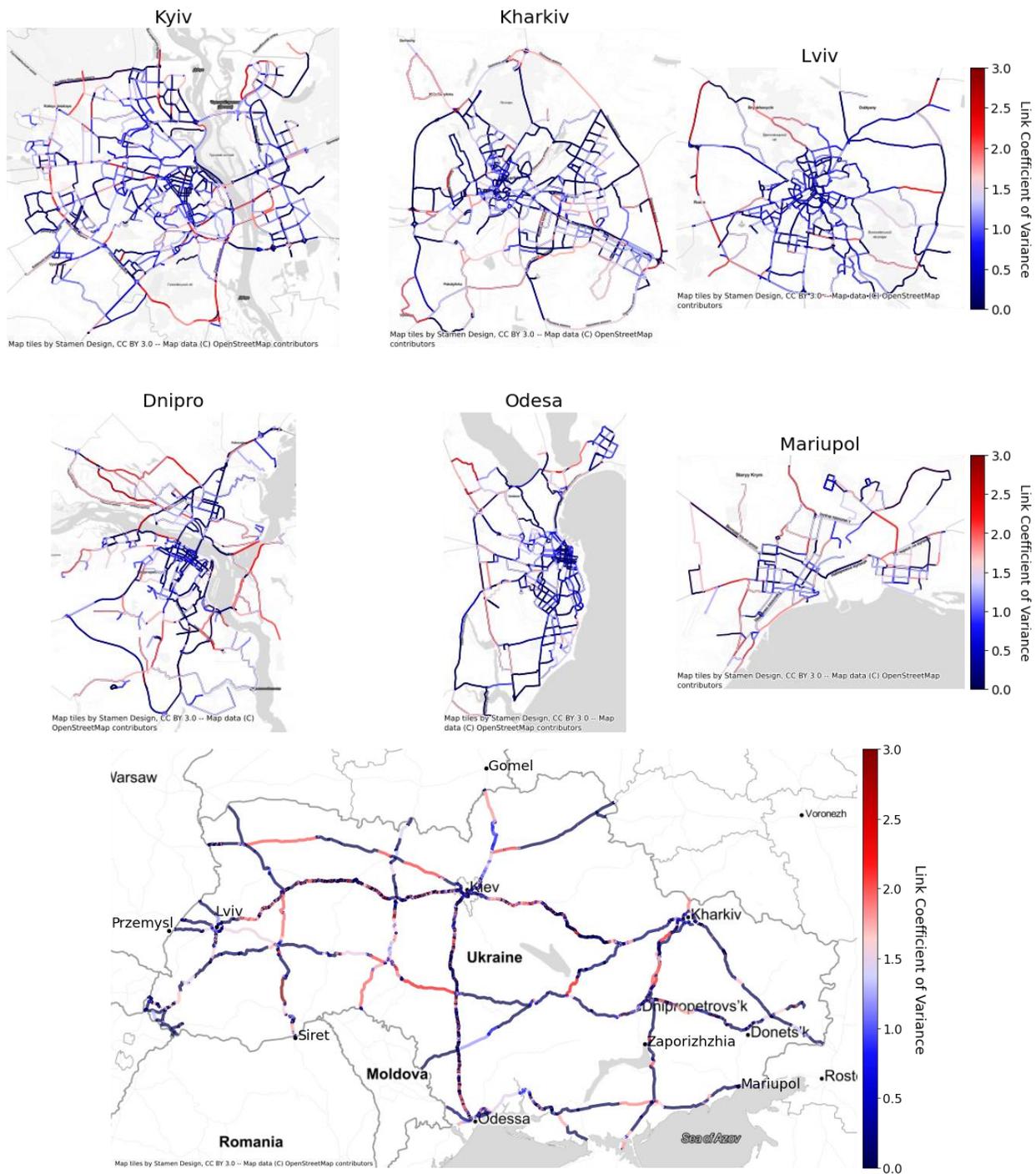

*Figure 2 Hue maps for link coefficient of variance over the whole study period*

**Temporal analysis of travel times variability and network congestion levels**

This section provides a temporal analysis of the effects of Ukraine war events using link travel time variability and network congestion levels. First, Figure 3 plots the temporal travel time variability using the network-wide average of the 7-day moving coefficient of variance (Eq. 2) for link travel times at three departure times. The metric helps in understanding the moving temporal variability in travel time uncertainty over a week period. Then Figure 4 plots the 7-day moving network-averaged link congestion indexes (Eq. 4). Each subplot from both figures visualizes three different time-of-day results from the morning, evening, and whole-day link travel times for each of the six cities focused in this study. Moreover, to help interpret the plot trends, we also label the reference numbers of the key war events that occurred in each city during the study period (listed in Table 1). Below, we analyze the results from each city separately, considering its situation and the timeline of events mentioned in section *'Timeline'*.

**Kyiv:** For Kyiv, two prominent trends are present in travel time variability (Figure 3), i) the times after curfew events 3 (February 26 to 28) and 7 (March 15 to 17) show a clear increase in travel time variability, ii) two period of times show a clear (decreasing) trend of calmness and regularity in congestion patterns, one between March 1 and March 14, in which less severe attacks on Kyiv are recorded and the other around the de-occupation time of Kyiv region (April 2) (decreasing trend for the morning and whole day travel time variability). Likewise, the only prominent trend visible in Figure 4 is around the Kyiv region de-occupation event (April 2 2022), where the morning congestion levels continuously decrease until the event weekend (i.e., the calmest morning of the study period), while the whole day congestion levels stay low for next few days.

**Kharkiv:** While Table 1 only mentions a few key war events for Kharkiv, it has been under consistent attack from Russia and shows a slight but continuous increasing trend in travel time variability (Figure 3) till event 7 (multiple attacks from Russian troops with up to 28 civilian casualties). However, after event 7, while the morning travel time variability remains rather high, overall, the travel time variability shows a continuous decreasing trend towards calmness and regularity in congestion patterns due to less severe attacks.

**Lviv:** Since Lviv is located in west Ukraine, farthest from Russia or its occupied territories, it has much fewer severe events and shows consistent travel time variability (Figure 3) apart in one period between March 26 and April 10. During this period, both travel time variability and congestion levels show higher trends for the evening and whole day times. These trends can be reasoned by i) the event 4 (two large missile attacks on Lviv) occurring on March 26, after which the higher uncertainty in travel time and congestion levels can be seen, and ii) the de-occupation of the Kyiv region (around April 2), causing the return of people from west of Ukraine and abroad, many of which went through Lviv acting as the major logistic and humanitarian hub.

**Mariupol:** Mariupol has been one of the prime occupation targets for Russia to create an overland corridor to occupied Crimea. Therefore, the city was practically under siege within the first week (February 28 saw the utilities cut off in most areas, and by March 1, Russia had blocked the city). The situation changed with the opening of a 'Green corridor' for evacuation (event 9, from March 15 to 18), during which the city showed a steep increase in travel time variability and congestion levels. Further, the occupation battles continue to occur (event 12) between March 23 to 28, maintaining similar increasing trends in both metrics. It is also essential to signify that by March 17 (event 11) already, half of the city was in Russian occupation, which by March 28 (end of event 12) increased to almost the whole city. Therefore, the increased variability trends in travel times and congestion levels after March 28 occurred under Russian occupation.

**Dnipro:** Apart from being constantly under attack, Dnipro also acted as an evacuation hub and home to thousands displaced by the war (being located near Kharkiv and Mariupol). Therefore, much of the variability in both travel times and congestion levels can be referred to the general war timeline. Overall, each of the two data collection periods shows prominent travel time variability peaks (Figure 3), i.e., between March 2 to 16 and March 27 to April 12. The variability trends during the first peak period can be attributed to war events in Kyiv, Kharkiv, and Mariupol, in which the evening times consistently show high values for both metrics. Similarly, the variability trends during the second peak period are mainly significant for morning time and can be attributed to events 3-5, i.e., regular rocket attacks in Dnipro between (March 30 to April 2). Note that the time frame also coincides with the de-occupation of the Kyiv region.

**Odesa:** For Odesa, multiple prominent variability trends are seen in link travel times and network congestion levels. First, both metrics show rapid increases after the curfew period from April 9 to 11, where the morning time congestion levels see a very sharp rise. Then, while the travel time variability calms until March 15, the large set of attacks by Russia (event 5) on March 15 and 16 retriggers its increase. Similarly, Figure 4 shows three different sets of congestion level peaks for morning or evening times, some of which can be related to the attacks on April 3 and 7 (events 6 and 7), which seem to trigger a sharp rise in congestion levels for the next day.

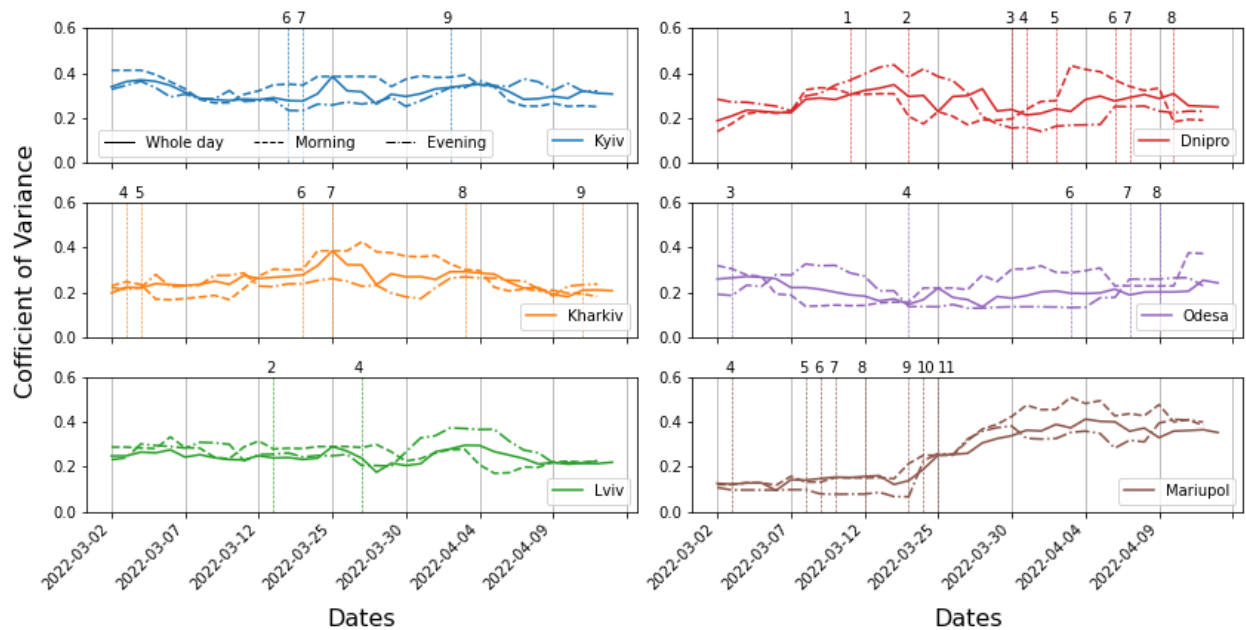

*Figure* **3** *Network averaged link coefficient of variance for travel times (7-day moving)*

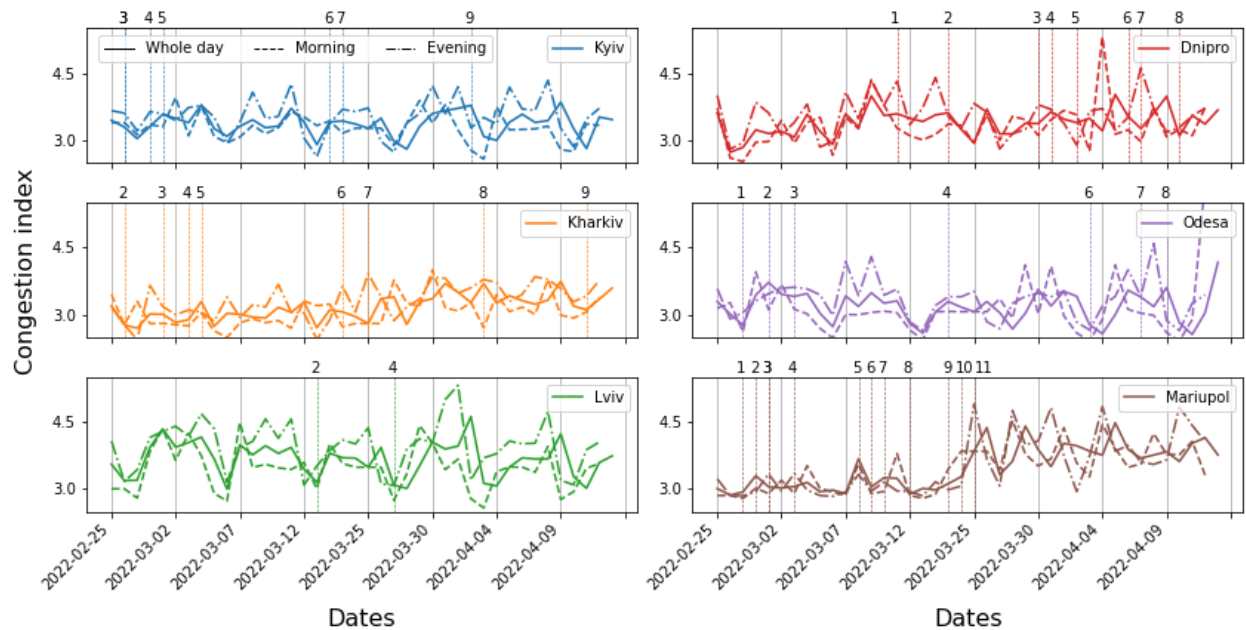

*Figure 4 Network averaged congestion index*

**ANALYSING TRAVEL PATTERNS BY OD DEMAND ESTIMATION**
Historically, researchers and practitioners depended on OD demand data acquired through Household Travel Surveys (HTS) (*38*). These surveys are conducted as one-time events every few years, are labor intensive, expensive, and frequently out of date (*18*). The HTS data are useful in long-term planning models, but they are not suitable for day-to-day traffic analysis or developing operational plans during disruptive events. Many investigations made use of non-traditional data sources such as loop detectors, Bluetooth, and registration plates (*20*). However, they lack spatial coverage, are time-consuming to collect, costly to set up and operate, and are also not suitable for rapid and automated planning in combat circumstances where data collecting, and the processing time are critical. Therefore, alternate methods are essential.

**Trip table estimation**
The previously noted OD estimation methodology is employed using TomTom travel time data on three different days. These three days correspond to i) the start of the invasion, ii) an intermediate time or critical event, and ii) the last day of the study period. We confine our investigation to three cities (Kyiv, Kharkiv and Mariupol) and three specific dates due to space constraints. We used a server with 40 cores, 3.1 GHz processing speed and a memory of 512 GB. We first defined the network zoning configuration, including the number of zones and centroids. Then, the software utilises link travel times and runs a customised genetic algorithm-based metaheuristic technique to derive OD matrices. These matrices were then used as source data for a network modelling process to estimate key metrics such as travel times, trip length, and congestion levels. This OD estimation technique can be seen as an advancement on past methods that rely on extensive and costly surveying of household travel behaviour, thus providing a viable option for developing countries facing data poverty challenges (*18*). Table 2 shows the percentage changes in key metrics compared to the base day, i.e., February 28 2022, for each of the three cities.

*Table 2 Key Statistics from the OD Estimation Analysis*

| City | Date | % change in average trip length compared to the base case | % change in average travel time compared to the base case | % change in total demand compared to the base case |
|---|---|---|---|---|
| Kyiv | February 28 2022 | - | - | - |
|  | March 16 2022 | -5.52 | -0.28 | +3.90 |
|  | April 12 2022 | +2.74 | +1.92 | +0.11 |
| Kharkiv | February 28 2022 | - | - | - |
|  | March 31 2022 | -3.14 | +1.55 | +6.05 |
|  | April 12 2022 | +3.40 | +11.79 | +2.63 |
| Mariupol | February 28 2022 | - | - | - |
|  | March 16 2022 | +13.11 | +28.44 | -2.50 |
|  | April 12 2022 | -6.76 | -11.66 | +0.58 |

**Zone production levels and destination congestion index**
To further examine the travel behavior patterns during the Ukraine invasion, this section analyses the zone averaged congestion indexes and demand generation metrics for the three cities of Kyiv, Kharkiv, and Mariupol. First, Figure 5 spatially visualizes the zonal destination congestion index to examine zone (attraction) impedance or accessibility, then Figure 6 spatially visualizes the zone production percentages to examine the demand contribution patterns. Below, we discuss the analysis of each city separately:

**Kyiv:** By February 28, Kyiv had got a series of airstrikes and rockets, a curfew, the Kyiv metro going into shelter mode, and the Russian military occupation up till near the town Bucha, i.e., northwest of Kyiv. However, for February 28, both metrics in Figures 5 and 6 mostly show normality, with high levels in the city's dense central area and low levels in almost all outer city regions. Most of the patterns continue for March 16 and April 12 because of Kyiv's strong defense forces failing all occupation attacks by Russia. However, two prominent trends are evident from both figures. First, a steep rise in congestion levels and trip productions are seen in the west city region on March 16, which indicates the advancements of the Russian military that by the time had occupied the outskirts regions (including Bucha) in northwest and west of Kyiv, resulting in people migrating towards Kyiv and increasing congestion levels on the west. Later, this trend is reversed back by April 12 due to the de-occupation of Kyiv. Then, the other trend of consistently higher levels of congestion in the southwest city area is also visible. The region mainly contains rural highways which connect Kyiv to the southwestern region, and the increase in congestion levels is arguably due to the network serving as a migration route since one of the two major highways connecting Kyiv to west Ukraine was occupied by Russia (northwestern highway through Bucha) while the other was disconnected by destroying a large bridge (western highway).

**Kharkiv:** Like Kyiv, by February 28, Kharkiv had already had many attacks from the Russian military, including tens of shelling attacks and even several Russian tanks entering the city (event 1-2). However, none of the Kharkiv city areas came under Russian occupation throughout the study period. Therefore, apart from the eastern, southeastern, and northwestern city areas, the patterns for zone productions and destination congestion index are rather consistent, i.e., high levels in the densely populated city central area and relatively lower levels in the outer city areas. The eastern city area of Saltivka saw consistent mass attacks by Russia (Table 1, *24*) and evidently saw a distributed increase in congestion levels, which interestingly also doesn't translate into an increase in production for that region. A similar trend is also seen for the northwestern suburban area of Derhachi, which also shows a sharp rise in congestion levels only for the April 12 plots without any increase in trip productions. This trend also results due to the consistent shelling of Russian troops on Derhachi between April 9 to 11 (*24*). Finally, the southeastern (industrial) region of Kharkiv, which was also heavily shelled by Russian tanks and artillery on April 10, also shows similar patterns of increase in congestion levels.

**Mariupol:** Mariupol city is administratively divided into four districts 'Kalmiuskyi' (northern city area), 'Livoberezhnyi' (left bank or southeastern city area), 'Prymorskyi' (southwestern city area containing the seaport) and 'Tsentralnyi' (the central city area in the middle-west). February 28 was a day before the blockade of Mariupol city by Russia. By that time, the city did not suffer many Russian attacks; therefore, the plots show close-to routine travel patterns, i.e., in Figure 5, the central city area and the port area show a higher destination congestion index, and in Figure 6, central city area, port area, and the west part of the left bank region show high production rates.

Then, March 16 is the second day of the three-day green evacuation corridor period meant to allow the citizens to evacuate the city. However, for congestion levels, only a slight increase is seen in the western and southwestern city regions, whereas the metric is unchanged for the northern city region, both of which should generally be considered evacuation routes. Further, on this day, the Russian army broke through the eastern part of the city (event 10), an event whose effect is visible because the congestion index levels show the activity with much higher levels in the left bank area. Concerning production rates, they significantly reduce for the central city area (less to no routine activities), slightly increase for the port (possible military movement) and remain similar but considerably distributed in the left bank and northeast regions (indicating Russian troops movement).

Finally, by April 12, most of the Mariupol city was under Russian occupation, and the zone destination congestion index levels were again back to lower level for the left bank, similar (distributed) for the central city and much higher for the northern city area. Note that the congestion patterns do have a significant change intending Russian military activities in the city since most Ukrainian soldiers and a lot of civilians were (by then) hiding in Azovstal plant (east of the port).

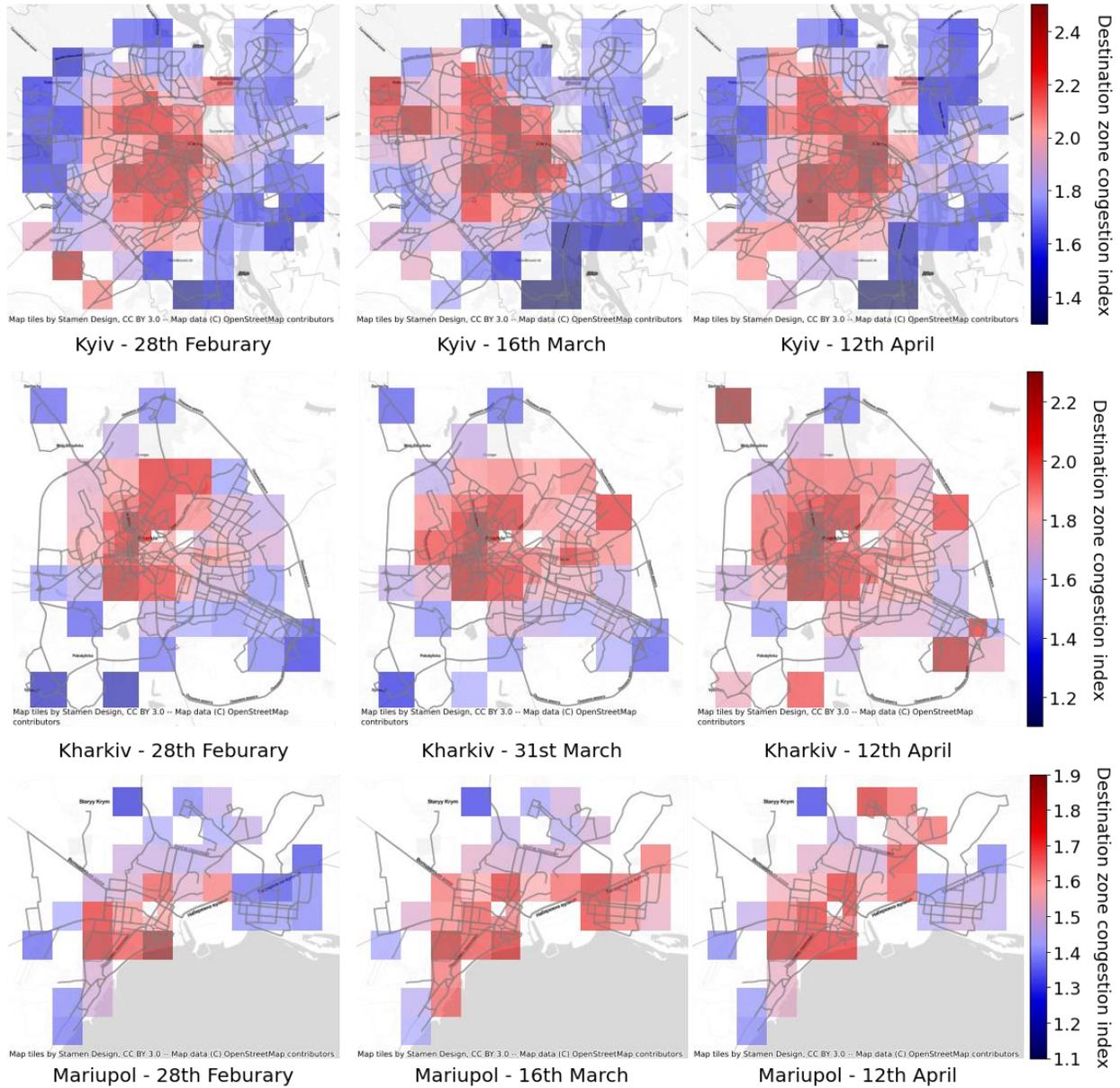

*Figure 5 Hue maps of average congestion index for zones as the destination*

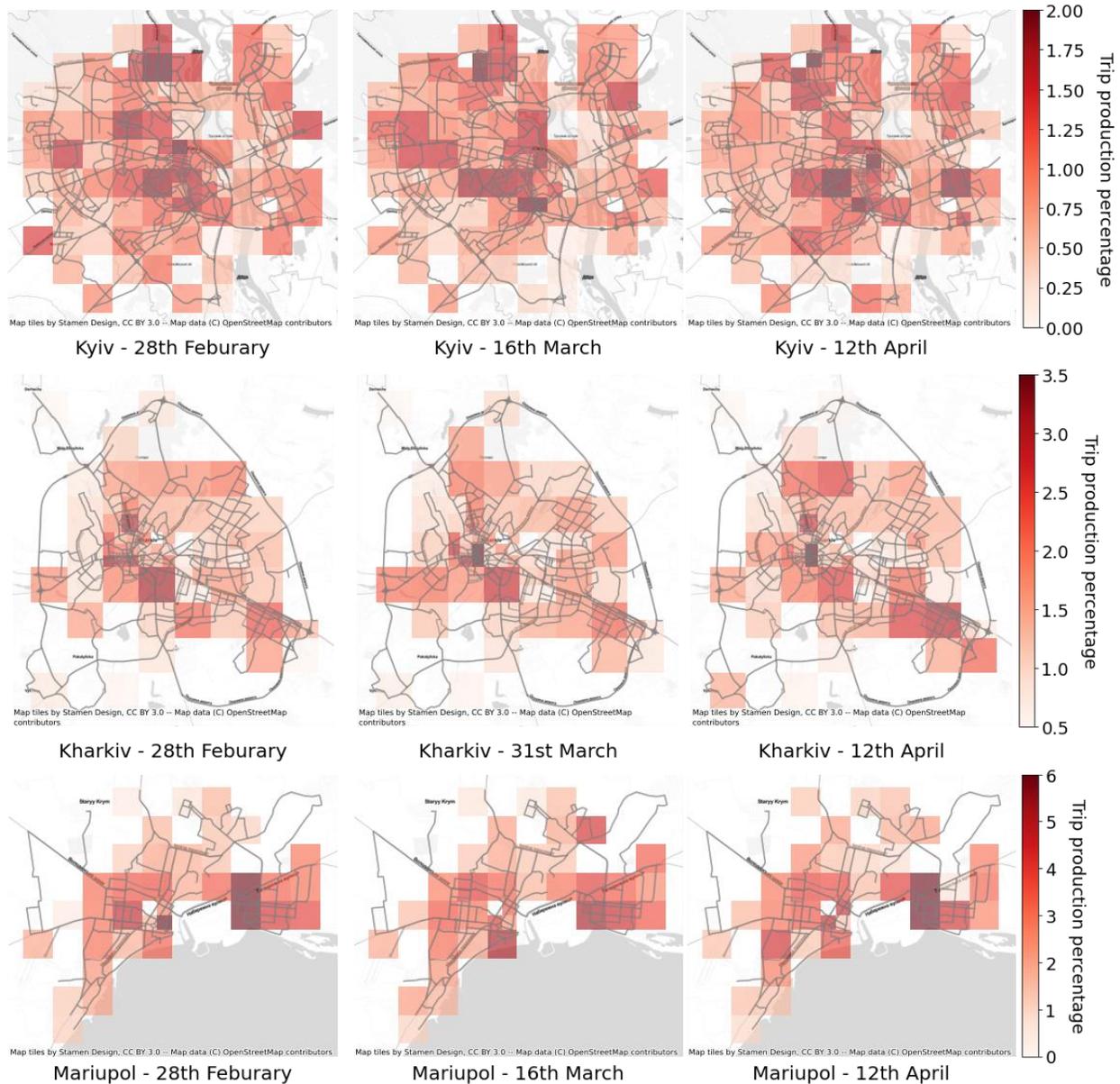

*Figure 6 Hue maps of percentage shares of zone productions*

## CONCLUSIONS

Disruptive events such as a national invasion are rare, and therefore disruption-induced travel patterns are not commonly analysed. The presented research utilised crowd-sourced pervasive traffic data from TomTom to evaluate traffic patterns and model origin-destination traffic demand during the ongoing Ukrainian conflict as it unfolded. The analysis demonstrated the utility of such data to determine and relate the impacts on travel patterns during such a major disruptive event.

Specifically, the research utilized and extended a recently developed demand estimation methodology to model and analyse travel impacts during the Ukraine invasion. The methodology enables researchers to quickly compare various policies and events by evaluating the trip tables across various networks and help

*Waller et al.*understand aggregate time-period-based demand changes or analyse the impacts of network and demand changes on critical performance metrics. Critically, the approach differs from basic analytics in that a network traffic assignment model is created, which allows comparative analysis within a planning approach. It should be noted, however, that the model produces highly aggregate outputs but is useful in situations where more traditional modelling solutions are not possible or prohibitive in terms of resourcing, particularly in rapidly evolving large-scale human conflict.

To conduct the study and relate the travel analysis, 51 key war event periods were synthesized (Table 1). For absolute temporal travel time variability, Mariupol shows the highest level at 0.5 under Russian occupation. Kyiv, Kharkiv, and Dnipro show up to a value of 0.4, whereas for the relative increase in variability, Mariupol, Dnipro, and Kharkiv show up to 200%, 135%, and 100% change during the study period. Finally, for the three majorly attacked cities of Kyiv, Kharkiv, and Mariupol, the invasion events triggered up to 50%, 55%, and 30% increase in congestion levels and up to 200%, 150%, and 300% demand productions, respectively.

One limitation of the current study is that due to the damage to the road network, parts of the network could be impaired. While the presented findings align at the highly aggregated level of analysis used in this study, future research will endeavour to model the destructive variant of the network design problem where system capacity will join the origin-destination demand values within the evolutionary algorithm variable set.

## ACKNOWLEDGMENTS
The authors acknowledge the financial support from the Chair of Transport Modeling and Simulation within the "Friedrich List" Faculty of Transport and Traffic Sciences at the Technical University of Dresden, Germany.
## AUTHOR CONTRIBUTIONS
Conceptualisation, S.T.W.; methodology, S.T.W., M.Q. and S.C.; software, M.Q. and S.C.; formal analysis, M.Q. and S.C.; investigation, S.T.W., M.Q., A.S., L.K. and S.C.; resources, S.T.W.; data curation, M.Q., A.S. and S.C.; writing—original draft preparation, M.Q., A.S., L.K. and S.C.; writing—review and editing, S.T.W., M.Q., A.S. and S.C.; visualisation, M.Q.; supervision, S.T.W.; project administration, S.T.W.; funding acquisition, S.T.W. All authors have read and agreed to the published version of the manuscript.